\documentclass[sigconf]{acmart}

\usepackage[T1]{fontenc}
\usepackage{xspace}
\usepackage{soul}
\usepackage{amsfonts}
\usepackage{url}
\usepackage{graphicx}
\usepackage{amsmath}
\usepackage{amsthm}
\usepackage{booktabs}
\usepackage[switch]{lineno}
\usepackage{hyperref}
\usepackage{cleveref}
\usepackage{multirow}
\usepackage{enumitem}
\usepackage{bbding}
\usepackage{comment}
\usepackage{subcaption}
\usepackage[noend]{algpseudocode}
\usepackage[T1]{fontenc}
\usepackage{flushend}
\usepackage[normalem]{ulem}
\usepackage{subfloat}
\usepackage{circledsteps}

\usepackage{geometry}
\usepackage{float}
\usepackage{caption}
\usepackage{bm}
\usepackage{makecell}
\usepackage{mathrsfs}
\usepackage{url}
\usepackage[linesnumbered,ruled,vlined]{algorithm2e}
\usepackage{color, xcolor}
\useunder{\uline}{\ul}{}

\newcommand{\method}{\textit{DRFormer}\xspace}

\AtBeginDocument{%
  }

\copyrightyear{2024} 
\acmYear{2024} 
\setcopyright{acmlicensed}\acmConference[CIKM '24]{Proceedings of the 33rd
ACM International Conference on Information and Knowledge
Management}{October 21--25, 2024}{Boise, ID, USA}
\acmBooktitle{Proceedings of the 33rd ACM International Conference on
Information and Knowledge Management (CIKM '24), October 21--25, 2024,
Boise, ID, USA}
\acmDOI{10.1145/3627673.3679724}
\acmISBN{979-8-4007-0436-9/24/10}
\begin{document}

\title{DRFormer: Multi-Scale Transformer Utilizing Diverse Receptive Fields for Long Time-Series Forecasting}

\author{Ruixin Ding}
\authornote{Both authors contributed equally to this research.}
\orcid{0009-0000-0462-8426}
\affiliation{%
  \institution{East China Normal University}
  \city{Shanghai}
  \country{China}
}
\email{51265901027@stu.ecnu.edu.cn}

\author{Yuqi Chen}
\authornotemark[1]
\orcid{0000-0003-4181-5794}
\affiliation{%
  \institution{Fudan University}
  \city{Shanghai}
  \country{China}
}
\email{chenyuqi21@m.fudan.edu.cn}

\author{Yu-Ting Lan}
\orcid{0000-0001-7132-5548}
\affiliation{%
  \institution{Shanghai Jiao Tong University}
  \city{Shanghai}
  \country{China}
}
\email{lanyuting8806@sjtu.edu.cn}

\author{Wei Zhang}
\authornote{Corresponding author.}
\orcid{0000-0001-6763-8146}
\affiliation{%
  \institution{East China Normal University}
  \city{Shanghai}
  \country{China}
}
\email{zhangwei.thu2011@gmail.com}

\renewcommand{\shortauthors}{Ruixin Ding, Yuqi Chen, Yu-Ting Lan, and Wei Zhang}
\renewcommand{\shorttitle}{DRFormer}
\begin{abstract}

  Long-term time series forecasting (LTSF) has been widely applied in finance, traffic prediction, and other domains. Recently, patch-based transformers have emerged as a promising approach, segmenting data into sub-level patches that serve as input tokens. However, existing methods mostly rely on predetermined patch lengths, necessitating expert knowledge and posing challenges in capturing diverse characteristics across various scales. Moreover, time series data exhibit diverse variations and fluctuations across different temporal scales, which traditional approaches struggle to model effectively. In this paper, we propose a dynamic tokenizer with a dynamic sparse learning algorithm to capture diverse receptive fields and sparse patterns of time series data. 
  In order to build hierarchical receptive fields, we develop a multi-scale Transformer model, coupled with multi-scale sequence extraction, capable of capturing multi-resolution features. Additionally, we introduce a group-aware rotary position encoding technique to enhance intra- and inter-group position awareness among representations across different temporal scales.
  Our proposed model, named \method, is evaluated on various real-world datasets, and experimental results demonstrate its superiority compared to existing methods. Our code is available at: \url{https://github.com/ruixindingECNU/DRFormer}.
  
\end{abstract}



\maketitle

\section{Introduction}

Time series forecasting is crucial in various domains such as finance~\cite{guo2021mrc,durairaj2022convolutional}, traffic prediction~\cite{zheng2020gman,zhang2021traffic,ji2023spatio,chen2023rntrajrec}, etc. The ability to accurately predict future values in time series data has significant implications for decision-making and planning~\cite{Wu2021AutoformerDT,zhou2021informer,chen2023contiformer}. The rapid advancement of deep learning has fueled remarkable progress in time series forecasting~\cite{jin2023time,chang2023llm4ts,gruver2023large}. Among various deep learning approaches, Transformer~\cite{Wu2021AutoformerDT,zhou2022fedformer,nie2022time,zhang2022crossformer} and MLP-based~\cite{zeng2023transformers,zhong2023multi} models have demonstrated superior performance due to their ability to capture long-term dependency. Furthermore, recent works have witnessed a significant breakthrough in patch-based transformers~\cite{triformer, nie2022time, shabani2023scaleformer} for the long-term time series forecasting (LTSF) task. These approaches divide time-series data into sub-level patches and utilize Transformer models~\cite{vaswani2017attention} to generate meaningful input features. However, existing methods are mostly designed to break the time series into patches of a fixed length~\cite{triformer, nie2022time} or with a set of predetermined patch lengths~\cite{shabani2023scaleformer}. 
This static patching with fixed patch length requires expert knowledge and poses challenges for extracting temporal features and dependencies from various scales of temporal intervals. 

To illustrate these challenges more comprehensively, it is essential to consider the following aspects: 
\textrm{(i)} The optimal sizes for patch division are influenced by the complex inherent characteristics and dynamic patterns of time series data, such as periodicity and trends. These intricate temporal patterns involve diverse variations and fluctuations across different temporal scales~\cite{chen2024multi}. Currently, no established rules exist that can be validated either experimentally or theoretically to determine the optimal patch length.
\textrm{(ii)}  Real-world time series usually present multi-periodicity, such as daily, weekly, and monthly variations for traffic conditions~\cite{Lai2017ModelingLA,hong2020heteta}, or weekly and quarterly variations for electricity consumption~\cite{wu2022timesnet}. These short-term and long-term recurring patterns contribute to the complexity of the forecasting task.
\textrm{(iii)}  The overall trend across the entire period and the specific time points of the learned sparse patterns are significant for the LTSF task. The morning and evening peaks typically offer crucial information for traffic prediction. These characteristics require careful model design to introduce proper inductive bias.


To address these challenges, we propose a novel dynamic patching strategy coupled with a group-aware Roformer~\cite{su2024roformer} network for LTSF. The proposed dynamic patching approach incorporates a dynamic sparse learning algorithm~\cite{xiao2022dynamic}, which overcomes the need for expert knowledge by learning diverse receptive fields and extracts sparse patterns to identify critical points, thereby making it more applicable to real-world scenarios. 
To capture the inherent multi-resolution features, we introduce a Transformer model that enables multiple scales of temporal modeling.
Additionally, we present a novel group-aware RoPE~\cite{su2024roformer} method, named gRoPE, to enhance intra- and inter-group position awareness among representations with different temporal scales. By incorporating group awareness, \method can effectively capture complex dependencies and interactions among different groups of representations, leading to improved forecasting performance. The contributions of the paper are as below:

\begin{itemize}[leftmargin=*]
    \item We propose a multi-scale Transformer model, named \method, which employs a dynamic tokenizer to learn diverse receptive fields and utilizes multi-scale sequence extraction to capture inherent multi-resolution features.
    \item We introduce a group-aware rotary position encoding technique for learning intra- and inter-group relative position embedding. With such a design, \method excels at capturing intricate dependencies among representations with distinct temporal scales.
    \item We conduct extensive experiments to demonstrate the superiority of \method over various baseline models in diverse real-world scenarios.
    
\end{itemize}

\section{Related Work}
In this section, we discuss the related studies from the following aspects: transformer for long-term time series forecasting, CNNs for time-series forecasting, and relative position embedding.

\subsection{Transformer for Long-term Time Series Forecasting}

The adoption of Transformer-based models has emerged as a promising approach for long-term time series forecasting~\cite{kitaev2020reformer,zhou2021informer,Wu2021AutoformerDT,zhou2022fedformer,liu2022non,li2023towards,zhang2023multi}. Among these models, Reformer~\cite{kitaev2020reformer} proposes locality-sensitive hashing attention for efficient and scalable sequence modeling. Informer~\cite{zhou2021informer} employs ProbSparse self-attention to extract important keys efficiently. 
Autoformer~\cite{Wu2021AutoformerDT} introduces a novel decomposition framework, along with an auto-correlation attention mechanism. FEDformer~\cite{zhou2022fedformer} utilizes Fourier transformation to model temporal characteristics and dynamics.
Patch-based transformers~\cite{triformer, nie2022time, shabani2023scaleformer}, dividing time-series data into sub-level patches, have yielded significant enhancements in forecasting accuracy and complexity reduction. 
However, existing methods mainly model time series within limited or fixed patch lengths, which necessitate expert knowledge to select the optimal patch lengths and pose challenges in capturing diverse characteristics across varying scales. Very recently, ~\cite{chen2024multi} developed a multi-scale Transformer that divides the time series into different temporal resolutions. However, it fails to learn a wide range of receptive fields given the selection of multiple patch lengths and static patching. In contrast, our proposed \method can adeptly learn from a wide range of receptive fields, capture both overarching trends and nuanced variations, and extract the inherent multi-resolution properties of the data.


\subsection{CNNs for Time-Series Forecasting}

In addition to Transformers, convolutional neural networks (CNNs) are highly regarded by researchers in the time-series community~\cite{he2016deep,simonyan2014very,dosovitskiy2020image,fang2023learning}. To enhance the generalization capabilities of time-series tasks and encompass diverse receptive fields, \cite{oord2016wavenet} proposed the use of dilated convolution kernels as a structure-based low bandpass filter. Moreover, OS-CNN~\cite{tang2020omni} introduced the Omni-Scale block (OS-block) for 1D-CNNs, enabling the model to learn a range of diverse receptive fields. Additionally, DSN~\cite{xiao2022dynamic} presented a dynamic sparse network that can adaptively cover various receptive fields without the need for extensive hyperparameter tuning. Drawing inspiration from CNNs, we integrate dynamic sparse networks and multi-scale modeling into the Transformer structure, enabling the model to leverage the advantages of CNNs.

\subsection{Relative Position Embedding}

In the realm of natural language processing, several approaches for relative position embedding (RPE) have been proposed ~\cite{dai-etal-2019-transformer,raffel2020exploring,shaw2018self,ke2021rethinking,su2024roformer}.
Among these, RoPE~\cite{su2024roformer} is a representative approach that encodes relative position by multiplying the context representations with a rotation matrix. 
Additionally, the adoption of RoPE has been widespread among large language models as a means to extend the context windows~\cite{chen2023extending,touvron2023llama}.
In this paper, we apply RoPE to enhance position awareness in the LTSF task. Besides, we propose group-aware rotary position embedding to encode intra- and inter-group relative position information into the attention mechanism, which is better suited for extracting multi-scale characteristics in time series data.



\begin{figure}[t]
    \centering
    \includegraphics[width=\linewidth]{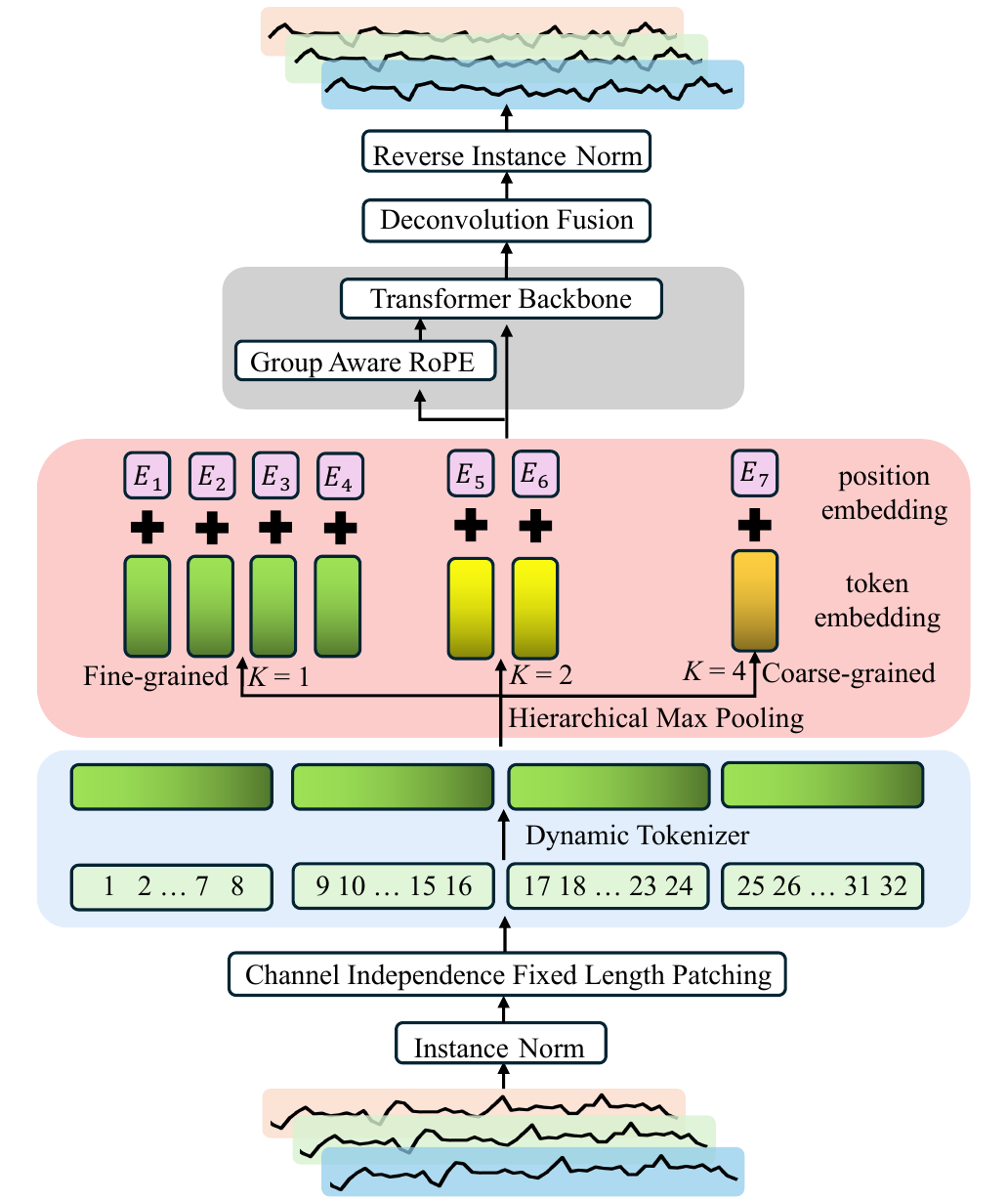}
    \caption{Overview of our \method. \method first utilizes a dynamic tokenizer to capture diverse receptive fields of each tokenizer. A hierarchical max pooling operation is then applied to leverage the multi-resolution property inherent in time series data. The multi-resolution time series data is then encoded by a group-aware Transformer model and finally processed by a deconvolution operation.}
    \label{fig:model}
\vspace{-0.1in}
\end{figure}

\section{Methodology}
In this section, we detailedly describe our method, \method, which captures multi-scale characteristics with diverse receptive fields and multi-resolution representations for LTSF, as shown in Figure~\ref{fig:model}.

We briefly introduce the intuition of our \method. As aforementioned, time series data is characterized by multi-scale properties and identifying critical timestamps provides crucial insights for prediction. To model such inductive bias, we propose a novel dynamic patching strategy coupled with a multi-scale Transformer to inject such characteristics priors into the forecasting pipelines. As illustrated in Figure 2, \method first incorporates a dynamic sparse network within the tokenizer, which simultaneously learns adaptive receptive fields and sparse patterns of the data. Next, we propose to transform time-series sequences into multi-scale sequences, allowing each token to represent features at multiple granularities. Finally, to capture cross-group interactions, we introduce a group-aware rotary position encoding technique for learning intra- and inter-group relative position embeddings. 

In the following, we first formulate the problem and give an overview of our method. Then, we delve into the details of the dynamic tokenizer, multi-scale sequence extraction and the multi-scale Transformer with group-aware rotary position embedding.

\subsection{Problem Formulation}
The task of time series forecasting involves predicting a future series of length\emph{-O} with the highest probability, based on a given past series of length\emph{-I}, denoted as \emph{input-I-predict-O}. In the context of long-term forecasting, the objective is to predict the future over a longer time horizon, specifically with a larger value of \emph{O}.  Given a multivariate time series, i.e., $\boldsymbol{X}^{1:I} \in \mathbb{R}^{I \times C}$, where $I$ denotes the length of the time series, and $C$ denotes the number of variates.  The general objective of this research is to predict $\boldsymbol{X}^{I+1:I+O} \in \mathbb{R}^{O \times C}$ with $\boldsymbol{X}^{1:I}$
as input. Note that our model is conducted on each variate of time series, i.e., channel independence. Thus, we denote $x_i$ as the time series for the $i$-th variate and omit the variate of time series for simplicity.


\subsection{Dynamic Tokenizer}

In this section, we describe how we discover and exploit various receptive fields adaptively with the dynamic sparse network. As aforementioned, static patching requires expertise to determine the length of temporal patch, in which complex inherent characteristics and dynamic patterns of time series data should be considered. Moreover, integrating fine-grained and coarse-grained features is crucial to model diverse variations and fluctuations, which is challenging for the pre-defined static model. To address this, we involve a novel dynamic tokenizer to dynamically capture the optimal scale features through a sparse learning strategy.


\begin{figure*}[t]
    \centering
    \setlength{\abovecaptionskip}{0.3cm}
    \includegraphics[width=\linewidth]{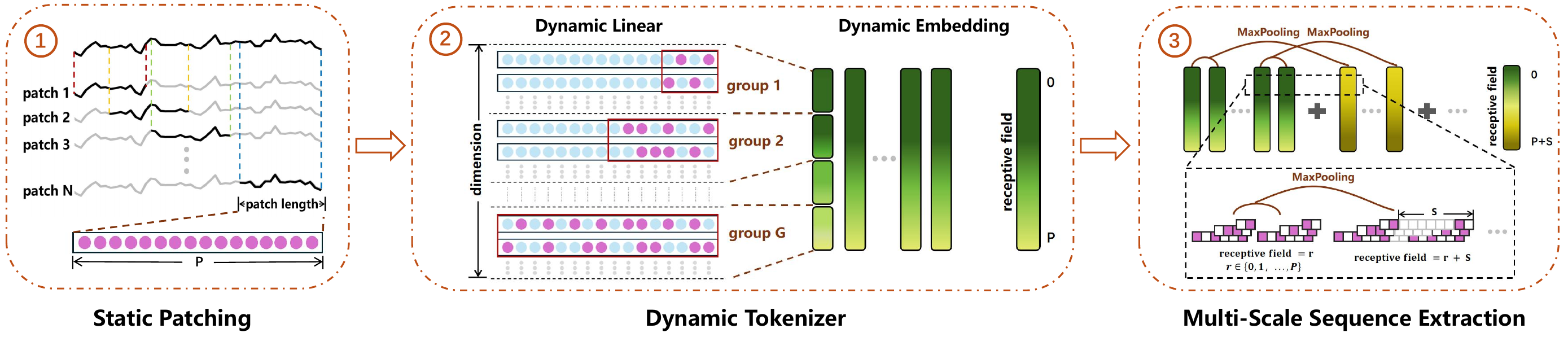}
    \caption{Illustration of static patching, dynamic tokenizer and multi-scale sequence extraction. \Circled{1} Taking $P$ = $16$ as an example, the input sequence is transformed into N patches. \Circled{2} The dynamic linear layer is divided into $G$ groups and the corresponding exploration regions for each group are shown in the red boxes. The number of group $G$ is set to $4$ and the sparse ratio $SR$ is set to $0.5$. Purple (blue) circles indicate activated (inactivated) weights. \Circled{3} Hierarchical max-pooling on patched tokens yields multi-group representations with a more comprehensive set of receptive fields as shown in Equation \ref{eq:multi-scale rf}.}
    \label{fig:dynamic}
\end{figure*}

\subsubsection{Data Normalization}

To migrate the distribution shift between the training and testing data~\cite{ulyanov2016instance,kim2021reversible,nie2022time}, we employ instance normalization on the input data. Specifically, we normalize each variable $x_i$ by subtracting its mean and dividing by its standard deviation before applying patching. After the output prediction, we add back the normalized values to restore the original distribution with the mean and standard deviation.

\subsubsection{Static Patching}

The proposed dynamic tokenizer, as depicted in Figure \ref{fig:dynamic}, first adopts a static patching operation to transform the input time series into sub-level patches~\cite{nie2022time}. Next, a dynamic linear transformation is applied to obtain dynamic representations, which serve as the input token embeddings. Specifically, we denote the patch length as $P$ and the stride as $S$. Each input univariate time series $x_i$ is first divided into patches $p_i \in \mathbb{R}^{P \times N}$ where $N$ is the number of patches, $N=\left\lfloor\frac{(I-P)}{S}\right\rfloor+2$. 

\subsubsection{Dynamic Linear Transformation}

Previous works adopt linear transformation to obtain input tokens~\cite{nie2022time}. Assume that $w^E \in \mathbb{R}^{P \times D}$ and $b^E \in \mathbb{R}^{D}$, where $D$ is the number of hidden dimensions of the model. The embeddings are obtained by $e_i=p_i^\top w^E+b^E$. However, these embeddings are limited by a fixed receptive field, i.e., all dimensions of each token have the same receptive field size of $P$. To address the limitation, we introduce a learnable sparse mask, i.e., 
\begin{equation}
f_i=p_i^\top (w^E \odot I(w^E))+b^E ~,
\end{equation}
where $I(\cdot): \mathbb{R}^{P \times D} \rightarrow \{0,1\}^{P \times D}$ denotes an indicator function~\cite{xiao2022dynamic}, $\odot$ denotes the element-wise product. A dynamic linear layer with sparse ratio $SR$ satisfies that $\|I(w^E)\|_0 \leq (1-SR) \times P \times D$. In Figure \ref{fig:dynamic}, a dynamic linear layer is depicted, showcasing the first, second, and last groups. The first dimension of each group learns receptive fields of sizes 3, 7, and 14, respectively. By definition of the token receptive field (tRF, as defined below), a dynamic linear layer is inherently designed to capture a comprehensive set of receptive fields, denoted as $RF = \{0, 1, ..., P\}$.

\noindent \textbf{Remark: Token Receptive Field (tRF) for dynamic linear layer.}
The receptive field (RF) in CNN layers is defined as the region in the input that the feature is looking at. In the context of the dynamic linear layer, tRF is defined as the region in the input that a token is looking at. 
Mathematically, assume that the indicator function of the weight vector $w^E_i$ is defined as $Ind=I(w^E_i) \in \{0,1\}^{P}$. Let $\mathcal{S}$ be the set of indices where $Ind_j =1 $, i.e., $\mathcal{S}=\{Ind_j=1 | 1 \leq j \leq P \}$, 
tRF is calculated as
\begin{equation}
tRF= 
\begin{cases} max(\mathcal{S}) - min(\mathcal{S}) +1, & \text { if } Ind \neq \textbf{0} \\ 0, & \text { otherwise }
\end{cases} ~.
\end{equation}

\subsubsection{Group Partition}

By design, during the training phase, the total number of activated weights must not exceed $(1-SR) \times P \times D$. However, a larger token receptive field occupies the majority of the tokens, especially as the sparsity ratio $SR$ decreases~\cite{xiao2022dynamic}, leading to a leak of local patterns being captured. To address this problem, we utilize a group partition strategy. In this approach, the dynamic linear layer is divided into several groups, whose corresponding exploration regions are of different sizes. Specifically, the weights $w^E \in \mathbb{R}^{P \times D}$ are split into $G$ groups along the output channel, that is,  $w_{1}^E, \cdots, w_{G}^E \in \mathbb{R}^{P \times \frac{D}{G}}$. For the $i$-th group, the exploration region comprises the last $\lceil \frac{iP}{G} \rceil$ positions, thereby ensuring that activated weights only appear within these positions and the number of activated weights must not exceed $(1-SR) \times \lceil \frac{iP}{G} \rceil \times \frac{D}{G}$. Additionally, we define the candidate set $\mathcal{C}$ as the set of weights that can be activated. Figure \ref{fig:dynamic} illustrates a dynamic tokenizer with four groups, where the first group allows activation only for the last four positions.

\subsubsection{Training the Indicator} 

\begin{algorithm}[htbp]
    \caption{Training algorithm for the indicator}
    \label{alg:1}
    \KwIn{Dataset $\mathcal{D}$, learning rate $\alpha$, initial weight $w^{E}$, candidate region $\mathcal{C}$.}
    \For{$t \leftarrow 1$ \KwTo $T$}{
        Sample a Batch $B_t \sim \mathcal{D}$\;
        $L_t = \sum_{i \in B_t} L\left(f_\theta\left(x_i\right), y_i\right)$\;
        Update $w^E$ and the network using gradient descent\;
        \If{$t \mod \Delta t = 0$}{
            Calculate $n$ using Eq. \eqref{eq:p} with $t, T, \alpha$ as inputs\;
            $\mathbb{I}_{\text{prune}} = \operatorname{ArgTopK}\left(-\left|w^E\right|, n\right)$\;
            $\mathbb{I}_{\text{grow}} = \operatorname{RandomK}\left(\mathcal{C} \cap [I(w^E) = 0], n\right)$\;
            $\mathbf{I}^l(.) \leftarrow \operatorname{Update} \mathbf{I}^l(.) \operatorname{using} \mathbb{I}_{\text{prune}} \operatorname{and} \mathbb{I}_{\text{grow}}$\;
        }
    }
\end{algorithm}

Updating the indicator directly through backpropagation is a non-differentiable operation. We adopt a heuristic algorithm to explore and update the weights~\cite{xiao2022dynamic}. In the selection of the masking strategy, various possibilities were explored. We ultimately determine masking out weights with small magnitudes as the masking strategy since it is intuitive and has been experimentally proven to be the most effective (more details in Section \ref{influence of strategy}). The whole algorithm is listed in Algorithm ~\ref{alg:1}. Specifically, assume we have a total of $T$ training iterations. For every $\Delta t$ iteration, we perform one step of update. At iteration $t$, since weights with smaller magnitudes contribute insignificantly or negligibly to the overall computation, we select $n$ weights with the smallest absolute values from the candidate set and set these weights to $0$, effectively deactivating them. To ensure recoverability from pruning, we randomly reintroduce $n$ weights, matching the number of pruned weights, to facilitate better exploration of activated weights. This dynamic and plastic weight exploration approach allows for adaptive exploration during the training process. The value of $n$ is controlled by the annealing function, which adjusts the pruning rate over time:
\begin{equation}
n=\frac{\alpha}{2}\left(1+\cos \left(\frac{t \pi}{T}\right)\right) \times \|I(w^E)\|_0 ~,
\label{eq:p}
\end{equation}
where $\alpha$ is a hyper-parameter to control the learning rate.

\subsection{Multi-Scale Sequence Extraction}

Time series data is characterized by both fine-grained local details and coarse-grained global composition, and capturing both aspects is crucial for comprehensive modeling. 
To address this, we propose a multi-scale approach that utilizes multi-group representations through hierarchical max-pooling on patched tokens. Specifically, we denote $f_i \in \mathbb{R}^{D \times N}$ as a latent representation of patching of the dynamic tokenizer. The hierarchical max-pooling strategy involves the application of max-pooling from fine-grained to coarse-grained with non-overlapping windows of diverse on consecutive patches to generate multi-resolution representations as follows:
\begin{equation}
\begin{aligned}
      \mathcal{F}_{i}^{0} & = \left\{f_{i}^{K_1}, f_{i}^{K_2}, \cdots, f_i^{K_k}\right\} 
\label{eq:input}
\end{aligned}
\end{equation}
where $k$ denotes the number of multi-scale sequences and $f_{i}^{K_1}$ is the original sequence from the dynamic tokenizer. Here, $f_{i}^{K_j} \in \mathbb{R}^{D \times \lceil \frac{N}{K_j} \rceil}, j \in \{1,2,\dots,k\}$ denotes the representation after max pooling operation, i.e.,
\begin{equation}
    f_{i,p}^{K_j} = \operatorname{MaxPooling}(f_{i,p}, f_{i,p+1},\ldots,f_{i,p+K_j-1}) ~,
\end{equation}
where $p+K_j-1 \leq N$ and $f_{i,p}$ denotes the $p$-th token in $f_i$. Besides, we denote $S_k=\{K_1, ..., K_k\}$ as the set of different kernels. As is shown in \Cref{fig:model}, we design the $S_k$ as a set of power two, i.e., $S_k=\{1, 2, ..., 2^{k-1}\}$, to empower the model with multi-scale ability and obtain a comprehensive representation from fine-grained to coarse-grained temporal information. 
With such a design, the tokens can capture a more comprehensive set of receptive fields of:
\begin{equation}
    \widehat{RF}=\{0, 1, ..., P+(2^{k-1}-1)\cdot S\} ~.
\label{eq:multi-scale rf}
\end{equation}

\subsection{Multi-Scale Transformers}

In this section, we formulate our multi-scale transformer. To overcome the limitations of position awareness of multi-scale representations for the transformer model, we propose a group-aware relative position encoding technique, which empowers our model to effectively capture intricate dependencies and interactions among different groups of representations, resulting in enhanced forecasting performance.

\subsubsection{Group-Aware Rotary Position Encoding}

Instead of using traditional absolute or relative position encoding, which ignores the inductive bias of intra and inter-group relations and treats different group embedding equally, we propose a novel group-aware rotary position encoding technique to capture intricate dependencies and interactions among different representation groups. We follow Roformer \cite{su2024roformer} and formulate the position encoding as the rotary matrix with pre-defined angle parameters. We derive the group-aware rotary position encoding for $f_{i}^{K_j} \in \mathbb{R}^{D \times \lceil \frac{N}{K_j} \rceil}$. Let $f_{i, m}^{K_j}$ be the $m$-th embedding for $i$-th group and the intra-group rotary position encoding for $f_{i, m}^{K_j}$ can be formulated as:
\begin{equation}
    \resizebox{\linewidth}{!}{$
    \boldsymbol{R}_{\Theta, i, m}^{d,\operatorname{intra}}=\left(\begin{array}{ccccccc}
    \cos \hat{m} \theta_1 & -\sin \hat{m} \theta_1 & 0 & 0 & \cdots & 0 & 0 \\
    \sin \hat{m} \theta_1 & \cos \hat{m} \theta_1 & 0 & 0 & \cdots & 0 & 0 \\
    0 & 0 & \cos \hat{m} \theta_2 & -\sin \hat{m} \theta_2 & \cdots & 0 & 0 \\
    0 & 0 & \sin \hat{m} \theta_2 & \cos \hat{m} \theta_2 & \cdots & 0 & 0 \\
    \vdots & \vdots & \vdots & \vdots & \ddots & \vdots & \vdots \\
    0 & 0 & 0 & 0 & \cdots & \cos \hat{m} \theta_{d / 2} & -\sin \hat{m} \theta_{d / 2} \\
    0 & 0 & 0 & 0 & \cdots & \sin \hat{m} \theta_{d / 2} & \cos \hat{m} \theta_{d / 2}
    \end{array}\right) ~,
    $}
\end{equation}
where $\hat{m}=m/\lceil \frac{N}{K_i}\rceil$, indicates the relative position of $m$ within the sequence, and pre-defined parameters:
\begin{equation}
    \Theta=\left\{\theta_i=10000^{-2(i-1) / d}, i \in[1,2, \ldots, d / 2]\right\} ~.
\end{equation}
Since $\boldsymbol{R}_{\Theta, i, m}^{d,intra}$ ignores the group information, we define another inter-group rotary position encoding as:
\begin{equation}
    \resizebox{\linewidth}{!}{$
    \boldsymbol{R}_{\Theta, i, m}^{d,\operatorname{inter}}=\left(\begin{array}{ccccccc}
    \cos i \theta_1 & -\sin i \theta_1 & 0 & 0 & \cdots & 0 & 0 \\
    \sin i \theta_1 & \cos i \theta_1 & 0 & 0 & \cdots & 0 & 0 \\
    0 & 0 & \cos i \theta_2 & -\sin i \theta_2 & \cdots & 0 & 0 \\
    0 & 0 & \sin i \theta_2 & \cos i \theta_2 & \cdots & 0 & 0 \\
    \vdots & \vdots & \vdots & \vdots & \ddots & \vdots & \vdots \\
    0 & 0 & 0 & 0 & \cdots & \cos i \theta_{d / 2} & -\sin i \theta_{d / 2} \\
    0 & 0 & 0 & 0 & \cdots & \sin i \theta_{d / 2} & \cos i \theta_{d / 2}
    \end{array}\right) ~.
    $}
\end{equation}
Here, intra- and inter-group rotary position encoding share the same parameters.

\subsubsection{Transformer Backbone}

The Transformer model is widely recognized for its effectiveness in sequence modeling tasks. However, to further improve its capacity to capture both inter-group and intra-group correlations, we introduce a novel group-aware rotary position encoding technique. Specifically, given the multi-scale inputs $\mathcal{F}_i^{l}$ and corresponding inter- and intra-group rotary position encoding, i.e., $\boldsymbol{R}_{\Theta}^{d,\operatorname{inter}}$, $\boldsymbol{R}_{\Theta}^{d,\operatorname{intra}}$. We first calculate keys and queries, i.e.,
\begin{equation}
    \{Q,K\}^{\{\operatorname{inter},\operatorname{intra}\}} = \mathcal{F}_i^{l}  \boldsymbol{W}_{\{Q, K\}} \boldsymbol{R}_{\Theta}^{d, \{\operatorname{inter},\operatorname{intra}\}} ~,
\end{equation}
where $\boldsymbol{W}_{\{Q, K\}}$ represents the transformation matrices for queries and keys, respectively. Next, we define the group-aware attention:
\begin{equation}
\operatorname{Attn}(\mathcal{F}_i^{l})=\operatorname{softmax}\left(\frac{\left(Q^{\operatorname{inter}} K^{\operatorname{inter}}\right)^\top + \left(Q^{\operatorname{intra}} K^{\operatorname{intra}}\right)^\top}{\sqrt{d_k}}\right) \mathcal{F}_i^{l}  \boldsymbol{W}_{V} ~.
\end{equation}
The multi-scale Transformer is a highly efficient model that effectively extracts multi-scale information from time series data while also capturing group awareness. Additionally, each  Transformer layer incorporates a feed-forward network and layer normalization~\cite{vaswani2017attention}. The mathematical formulations are as follows:
\begin{equation}
\begin{aligned}
\mathcal{F}^{l,1}&=\mathcal{F}^{l-1} + \operatorname{LN}\left( \operatorname{Attn}(\mathcal{F}_i^{l-1}) \right) ~, \\
\mathcal{F}^{l}&=\mathcal{F}^{l,1}+ \operatorname{LN}\left( \operatorname{FFN}(\mathcal{F}_i^{l,1}) \right) ~, \\
\end{aligned}
\end{equation}
where $\mathcal{F}^{l}$ indicates the output for the $l$-th Transformer layer with the input $\mathcal{F}^{0}$ defined in Eq. \eqref{eq:input}, $\operatorname{LN}$ and $\operatorname{FFN}$ represent the layer normalization operation and the feed-forward network, respectively~\cite{vaswani2017attention}.

\subsubsection{Representation Fusion with Deconvolution}

One possible approach is to use these embeddings for prediction directly. However, to achieve predictions that incorporate both fine-grained local details and coarse-grained global composition, we propose a fusion technique that combines these representations using deconvolution operations. Specifically, the output from the Transformer backbone is first split into multi-scale sequences:
\begin{equation}
    o_i^m=\left\{o_i^{K_1}, o_i^{K_2}, \ldots, o_i^{K_k}\right\} ~.
\end{equation}
We then perform a deconvolution operation~\cite{xu2014deep}, which is a technique that upsamples features: 
\begin{equation}
de_i^{K_j} = \operatorname{Deconv}\left( o_i^{K_j}, K_j \right) ~,
\end{equation}
where $de_i^{K_j} \in \mathbb{R}^{N \times d}$. Finally, the output is obtained by:
\begin{equation}
O_i = \sum_{j=1}^{k} de_i^{K_j} ~.
\end{equation}

\subsection{Loss Function}
We adopt the Mean Squared Error (MSE) loss to measure the discrepancy between the forecasting results and the ground truth observations. Let $\hat{\boldsymbol{X}}^{I+1: I+O}$ and $\boldsymbol{X}^{I+1: I+O}$ be the predictions and real observations from time $I+1$ to $I+O$. We denote $\hat{\boldsymbol{x}}_i^{I+1:I+O}$ and $\boldsymbol{x}_i^{I+1: I+O}$ be the predictions and real observations from the $i$-th variate. The training loss is defined as:
\begin{equation}
\mathcal{L}=\mathbb{E}_{\boldsymbol{X} \sim \mathcal{D}} \left[ \frac{1}{C} \sum_{i=1}^{C} \left\|\hat{\boldsymbol{x}}_i^{I+1: I+O}-\boldsymbol{x}_i^{I+1: I+O}\right\|_2^2 \right] .
\end{equation}

\section{Experiments}


\begin{table*}[t]
  \centering
  \caption{Experimental results for multivariate time series forecasting. Bold (Underlined) values indicate the best (second-best) performance. The input length is $96$ for each dataset, and the prediction lengths for the ECL, Traffic, and ETT datasets are $\{ 96,192,336,720 \}$, while $\{24,36,48,60\}$ for the ILI dataset. (avg for the averaged results on the four different prediction lengths)}
  \resizebox{\textwidth}{!}{%
  \begin{tabular}{c|c|cc|cc|cc|cc|cc|cc|cc|cc|cc}
  \toprule
  \multicolumn{2}{c|}{Models} & \multicolumn{2}{c|}{\method} & \multicolumn{2}{c|}{Koopa} & \multicolumn{2}{c|}{PatchTST} & \multicolumn{2}{c|}{TimesNet} & \multicolumn{2}{c|}{Dlinear} & \multicolumn{2}{c|}{ETSformer} & \multicolumn{2}{c|}{Autoformer} & \multicolumn{2}{c|}{Informer} & \multicolumn{2}{c}{Reformer} \\
  \midrule
  \multicolumn{2}{c|}{Metric} & MSE & MAE & MSE & MAE & MSE & MAE & MSE & MAE & MSE & MAE & MSE & MAE & MSE & MAE & MSE & MAE & MSE & MAE \\
  \midrule
  \multirow{5}{*}{\rotatebox{90}{ECL}} & 96 & {\ul 0.163} & {\ul 0.254} & \textbf{0.147} & \textbf{0.247} & 0.178 & 0.264 & 0.168 & 0.272 & 0.197 & 0.282 & 0.187 & 0.304 & 0.201 & 0.317 & 0.274 & 0.368 & 0.312 & 0.402 \\
 & 192 & \textbf{0.174} & \textbf{0.264} & {\ul 0.181} & 0.276 & 0.184 & {\ul 0.270} & 0.184 & 0.289 & 0.196 & 0.285 & 0.199 & 0.315 & 0.222 & 0.334 & 0.296 & 0.386 & 0.348 & 0.433 \\
 & 336 & \textbf{0.193} & \textbf{0.282} & {\ul 0.195} & 0.290 & 0.201 & {\ul 0.286} & 0.198 & 0.300 & 0.209 & 0.301 & 0.212 & 0.329 & 0.231 & 0.338 & 0.300 & 0.394 & 0.350 & 0.433 \\
 & 720 & 0.232 & {\ul 0.317} & {\ul 0.229} & \textbf{0.316} & 0.241 & 0.319 & \textbf{0.220} & 0.320 & 0.245 & 0.333 & 0.233 & 0.345 & 0.254 & 0.361 & 0.373 & 0.439 & 0.340 & 0.420 \\
 & \textbf{avg} & {\ul 0.191} & \textbf{0.279} & \textbf{0.188} & {\ul 0.282} & 0.201 & 0.285 & 0.193 & 0.295 & 0.212 & 0.300 & 0.208 & 0.323 & 0.227 & 0.338 & 0.311 & 0.397 & 0.338 & 0.422 \\
\midrule
\multirow{5}{*}{\rotatebox{90}{Traffic}} & 96 & \textbf{0.414} & \textbf{0.267} & 0.477 & 0.317 & {\ul 0.454} & {\ul 0.290} & 0.593 & 0.321 & 0.650 & 0.396 & 0.607 & 0.392 & 0.613 & 0.388 & 0.719 & 0.391 & 0.732 & 0.423 \\
 & 192 & \textbf{0.427} & \textbf{0.271} & 0.500 & 0.339 & {\ul 0.461} & {\ul 0.291} & 0.617 & 0.336 & 0.598 & 0.370 & 0.621 & 0.399 & 0.616 & 0.382 & 0.696 & 0.379 & 0.733 & 0.420 \\
 & 336 & \textbf{0.440} & \textbf{0.278} & 0.531 & 0.349 & {\ul 0.477} & {\ul 0.299} & 0.629 & 0.336 & 0.605 & 0.373 & 0.622 & 0.396 & 0.622 & 0.337 & 0.777 & 0.420 & 0.742 & 0.420 \\
 & 720 & \textbf{0.474} & \textbf{0.296} & 0.566 & 0.366 & {\ul 0.510} & {\ul 0.316} & 0.640 & 0.350 & 0.645 & 0.394 & 0.632 & 0.396 & 0.660 & 0.408 & 0.864 & 0.472 & 0.755 & 0.423 \\
 & \textbf{avg} & \textbf{0.439} & \textbf{0.278} & 0.519 & 0.343 & {\ul 0.476} & {\ul 0.299} & 0.620 & 0.336 & 0.625 & 0.383 & 0.621 & 0.396 & 0.628 & 0.379 & 0.764 & 0.416 & 0.741 & 0.422 \\
   \midrule
  \multirow{5}{*}{\rotatebox{90}{ETTh1}} & 96 & \textbf{0.378} & \textbf{0.398} & {\ul 0.384} & 0.407 & 0.396 & 0.408 & {\ul 0.384} & 0.402 & 0.386 & {\ul 0.400} & 0.494 & 0.479 & 0.449 & 0.459 & 0.865 & 0.713 & 0.837 & 0.728 \\
 & 192 & \textbf{0.425} & \textbf{0.429} & 0.447 & 0.435 & 0.445 & 0.440 & {\ul 0.436} & 0.429 & 0.437 & {\ul 0.432} & 0.538 & 0.504 & 0.500 & 0.482 & 1.008 & 0.792 & 0.923 & 0.766 \\
 & 336 & \textbf{0.467} & \textbf{0.453} & 0.493 & 0.461 & 0.486 & 0.464 & 0.491 & 0.469 & {\ul 0.481} & {\ul 0.459} & 0.574 & 0.521 & 0.521 & 0.496 & 1.107 & 0.809 & 1.097 & 0.835 \\
 & 720 & \textbf{0.491} & \textbf{0.478} & {\ul 0.512} & {\ul 0.487} & 0.491 & 0.490 & 0.521 & 0.500 & 0.519 & 0.516 & 0.562 & 0.535 & 0.514 & 0.512 & 1.181 & 0.865 & 1.257 & 0.889 \\
 & \textbf{avg} & \textbf{0.440} & \textbf{0.440} & 0.459 & {\ul 0.448} & {\ul 0.455} & 0.451 & 0.458 & 0.450 & 0.456 & 0.452 & 0.542 & 0.510 & 0.496 & 0.487 & 1.040 & 0.795 & 1.029 & 0.805 \\
   \midrule
  \multirow{5}{*}{\rotatebox{90}{ETTh2}} & 96 & \textbf{0.290} & \textbf{0.345} & 0.314 & 0.357 & {\ul 0.298} & {\ul 0.347} & 0.340 & 0.374 & 0.333 & 0.387 & 0.340 & 0.391 & 0.346 & 0.388 & 3.755 & 1.525 & 2.626 & 1.317 \\
 & 192 & \textbf{0.367} & \textbf{0.393} & {\ul 0.378} & 0.398 & 0.382 & {\ul 0.396} & 0.402 & 0.414 & 0.477 & 0.476 & 0.430 & 0.439 & 0.456 & 0.452 & 5.602 & 1.931 & 11.120 & 2.979 \\
 & 336 & \textbf{0.414} & \textbf{0.427} & {\ul 0.419} & 0.491 & 0.420 & {\ul 0.431} & 0.452 & 0.452 & 0.594 & 0.541 & 0.485 & 0.497 & 0.482 & 0.486 & 2.723 & 1.340 & 4.028 & 1.688 \\
 & 720 & \textbf{0.426} & \textbf{0.446} & 0.445 & 0.456 & {\ul 0.433} & {\ul 0.449} & 0.462 & 0.468 & 0.831 & 0.657 & 0.500 & 0.497 & 0.515 & 0.511 & 3.467 & 1.473 & 5.381 & 2.015 \\
 & \textbf{avg} & \textbf{0.374} & \textbf{0.403} & 0.389 & 0.426 & {\ul 0.383} & {\ul 0.406} & 0.414 & 0.427 & 0.559 & 0.515 & 0.439 & 0.456 & 0.450 & 0.459 & 3.887 & 1.567 & 5.789 & 2.000 \\
   \midrule
  \multirow{5}{*}{\rotatebox{90}{ETTm1}} & 96 & \textbf{0.328} & {\ul 0.368} & {\ul 0.330} & \textbf{0.363} & 0.355 & 0.383 & 0.338 & 0.375 & 0.345 & 0.372 & 0.375 & 0.398 & 0.505 & 0.475 & 0.672 & 0.571 & 0.538 & 0.528 \\
 & 192 & \textbf{0.364} & \textbf{0.387} & 0.379 & 0.393 & 0.393 & 0.400 & {\ul 0.374} & \textbf{0.387} & 0.380 & 0.389 & 0.408 & 0.410 & 0.553 & 0.496 & 0.795 & 0.669 & 0.658 & 0.592 \\
 & 336 & \textbf{0.390} & \textbf{0.405} & {\ul 0.402} & 0.412 & 0.424 & 0.417 & 0.410 & {\ul 0.411} & 0.413 & 0.413 & 0.435 & 0.428 & 0.621 & 0.537 & 1.212 & 0.871 & 0.898 & 0.721 \\
 & 720 & \textbf{0.449} & \textbf{0.439} & 0.475 & 0.448 & 0.477 & {\ul 0.446} & 0.478 & 0.450 & {\ul 0.474} & 0.453 & 0.499 & 0.462 & 0.670 & 0.561 & 1.166 & 0.823 & 1.102 & 0.841 \\
 & \textbf{avg} & \textbf{0.383} & \textbf{0.400} & {\ul 0.397} & {\ul 0.404} & 0.412 & 0.412 & 0.400 & 0.406 & 0.403 & 0.407 & 0.429 & 0.425 & 0.587 & 0.517 & 0.961 & 0.734 & 0.799 & 0.671 \\
   \midrule
  \multirow{5}{*}{\rotatebox{90}{ETTm2}} & 96 & \textbf{0.175} & \textbf{0.259} & {\ul 0.179} & {\ul 0.261} & {\ul 0.179} & 0.263 & 0.187 & 0.267 & 0.193 & 0.292 & 0.189 & 0.280 & 0.255 & 0.339 & 0.365 & 0.453 & 0.658 & 0.619 \\
 & 192 & \textbf{0.241} & {\ul 0.303} & 0.245 & 0.306 & {\ul 0.244} & \textbf{0.302} & 0.249 & 0.309 & 0.284 & 0.362 & 0.253 & 0.319 & 0.281 & 0.340 & 0.533 & 0.563 & 1.078 & 0.827 \\
 & 336 & 0.305 & \textbf{0.345} & \textbf{0.304} & \textbf{0.345} & \textbf{0.304} & \textbf{0.345} & 0.321 & 0.351 & 0.369 & 0.427 & 0.314 & 0.357 & 0.339 & 0.372 & 1.363 & 0.887 & 1.549 & 0.972 \\
 & 720 & {\ul 0.408} & \textbf{0.400} & \textbf{0.406} & {\ul 0.402} & {\ul 0.408} & 0.405 & {\ul 0.408} & 0.403 & 0.554 & 0.522 & 0.414 & 0.413 & 0.433 & 0.432 & 3.379 & 1.388 & 2.631 & 1.242 \\
 & \textbf{avg} & \textbf{0.282} & \textbf{0.327} & {\ul 0.284} & {\ul 0.329} & {\ul 0.284} & {\ul 0.329} & 0.291 & 0.333 & 0.350 & 0.401 & 0.293 & 0.342 & 0.327 & 0.371 & 1.410 & 0.823 & 1.479 & 0.915 \\
   \midrule
  \multirow{5}{*}{\rotatebox{90}{ILI}} & 24 & \textbf{1.887} & \textbf{0.863} & {\ul 1.974} & 0.900 & 2.101 & {\ul 0.866} & 2.317 & 0.934 & 2.398 & 1.040 & 2.527 & 1.020 & 3.483 & 1.287 & 5.764 & 1.677 & 4.400 & 1.382 \\
 & 36 & {\ul 2.117} & \textbf{0.892} & 2.278 & 0.928 & 2.647 & 0.978 & \textbf{1.972} & {\ul 0.920} & 2.646 & 1.088 & 2.615 & 1.007 & 3.103 & 1.148 & 4.755 & 1.467 & 4.783 & 1.448 \\
 & 48 & \textbf{1.929} & \textbf{0.827} & 2.156 & 0.910 & {\ul 2.056} & {\ul 0.882} & 2.238 & 0.940 & 2.614 & 1.086 & 2.359 & 0.972 & 2.669 & 1.085 & 4.763 & 1.469 & 4.832 & 1.465 \\
 & 60 & \textbf{1.903} & \textbf{0.906} & 2.267 & 0.956 & 2.329 & 1.020 & {\ul 2.027} & {\ul 0.928} & 2.804 & 1.146 & 2.487 & 1.016 & 2.770 & 1.125 & 5.264 & 1.564 & 4.882 & 1.483 \\
 & \textbf{avg} & \textbf{1.959} & \textbf{0.872} & 2.169 & {\ul 0.924} & 2.283 & 0.937 & {\ul 2.139} & 0.931 & 2.616 & 1.090 & 2.497 & 1.004 & 3.006 & 1.161 & 5.137 & 1.544 & 4.724 & 1.445 \\
    \bottomrule
  \end{tabular}%
  }
  \label{tab:main}
  \end{table*}

\begin{table*}[t]
\centering
\caption{Experimental results for univariate time series forecasting. Bold (Underlined) values indicate the best (second-best) performance. The input length is $96$ for each dataset, and the prediction lengths for ETTh1, ETTh2, ETTm1, and ETTm2 datasets are $\{ 96,192,336,720 \}$, while $\{24,36,48,60\}$ for the ILI dataset. (avg for the averaged results on four different prediction lengths)}
\resizebox{\textwidth}{!}{%
\begin{tabular}{c|c|cc|cc|cc|cc|cc|cc|cc|cc|cc}
\toprule
\multicolumn{2}{c|}{Methods} & \multicolumn{2}{c|}{DRFormer} & \multicolumn{2}{c|}{Koopa} & \multicolumn{2}{c|}{PatchTST} & \multicolumn{2}{c|}{TimesNet} & \multicolumn{2}{c|}{FEDformer} & \multicolumn{2}{c|}{ETSformer} & \multicolumn{2}{c|}{Autoformer} & \multicolumn{2}{c|}{Informer} & \multicolumn{2}{c}{Reformer} \\
\midrule
\multicolumn{2}{c|}{Metric} & MSE & MAE & MSE & MAE & MSE & MAE & MSE & MAE & MSE & MAE & MSE & MAE & MSE & MAE & MSE & MAE & MSE & MAE \\
\midrule
\multirow{5}{*}{\rotatebox{90}{ETTh1}} & 96 & \textbf{0.056} & {\ul 0.183} & 0.058 & 0.184 & {\ul 0.057} & \textbf{0.179} & 0.058 & 0.185 & 0.079 & 0.215 & 0.063 & 0.194 & 0.071 & 0.206 & 0.193 & 0.377 & 0.532 & 0.569 \\
 & 192 & \textbf{0.072} & \textbf{0.206} & {\ul 0.073} & {\ul 0.207} & 0.075 & 0.209 & 0.077 & 0.213 & 0.104 & 0.245 & 0.085 & 0.227 & 0.114 & 0.262 & 0.217 & 0.395 & 0.568 & 0.575 \\
 & 336 & \textbf{0.084} & \textbf{0.230} & 0.089 & {\ul 0.232} & 0.089 & 0.233 & {\ul 0.088} & {\ul 0.232} & 0.119 & 0.270 & 0.100 & 0.251 & 0.107 & 0.258 & 0.202 & 0.381 & 0.635 & 0.589 \\
 & 720 & \textbf{0.088} & \textbf{0.233} & {\ul 0.094} & {\ul 0.241} & 0.097 & 0.245 & 0.095 & 0.242 & 0.142 & 0.299 & 0.100 & 0.250 & 0.126 & 0.283 & 0.183 & 0.355 & 0.762 & 0.666 \\
 & \textbf{avg} & \textbf{0.075} & \textbf{0.213} & {\ul 0.079} & {\ul 0.216} & 0.080 & 0.217 & 0.080 & 0.218 & 0.111 & 0.257 & 0.087 & 0.231 & 0.105 & 0.252 & 0.199 & 0.377 & 0.624 & 0.600 \\
 \midrule
\multirow{5}{*}{\rotatebox{90}{ETTh2}} & 96 & {\ul 0.133} & {\ul 0.279} & 0.139 & 0.290 & 0.137 & 0.285 & 0.141 & 0.293 & \textbf{0.128} & \textbf{0.271} & 0.157 & 0.310 & 0.153 & 0.306 & 0.213 & 0.373 & 1.411 & 0.838 \\
 & 192 & \textbf{0.183} & {\ul 0.336} & \textbf{0.183} & {\ul 0.336} & 0.187 & 0.340 & 0.193 & 0.347 & 0.185 & \textbf{0.330} & 0.211 & 0.364 & 0.204 & 0.351 & 0.227 & 0.387 & 5.658 & 1.671 \\
 & 336 & \textbf{0.218} & \textbf{0.373} & {\ul 0.219} & 0.385 & 0.223 & {\ul 0.378} & 0.233 & 0.386 & 0.231 & {\ul 0.378} & 0.258 & 0.409 & 0.246 & 0.389 & 0.242 & 0.401 & 4.777 & 1.582 \\
 & 720 & \textbf{0.227} & \textbf{0.383} & {\ul 0.238} & {\ul 0.393} & 0.244 & 0.397 & 0.255 & 0.405 & 0.278 & 0.420 & 0.289 & 0.434 & 0.268 & 0.409 & 0.291 & 0.439 & 2.042 & 1.039 \\
 & \textbf{avg} & \textbf{0.190} & \textbf{0.343} & {\ul 0.195} & 0.351 & 0.198 & 0.350 & 0.206 & 0.358 & 0.206 & {\ul 0.350} & 0.229 & 0.379 & 0.218 & 0.364 & 0.243 & 0.400 & 3.472 & 1.283 \\
 \midrule
\multirow{5}{*}{\rotatebox{90}{ETTm1}} & 96 & \textbf{0.028} & \textbf{0.125} & {\ul 0.030} & 0.130 & 0.031 & {\ul 0.127} & {\ul 0.030} & 0.129 & 0.033 & 0.140 & 0.032 & 0.135 & 0.056 & 0.183 & 0.109 & 0.277 & 0.296 & 0.355 \\
 & 192 & \textbf{0.043} & \textbf{0.159} & 0.045 & {\ul 0.161} & {\ul 0.044} & 0.165 & 0.047 & 0.163 & 0.058 & 0.186 & 0.046 & 0.167 & 0.081 & 0.216 & 0.151 & 0.310 & 0.429 & 0.474 \\
 & 336 & \textbf{0.057} & \textbf{0.184} & {\ul 0.060} & 0.188 & {\ul 0.060} & {\ul 0.187} & 0.063 & 0.193 & 0.084 & 0.231 & {\ul 0.060} & 0.188 & 0.076 & 0.218 & 0.427 & 0.591 & 0.585 & 0.583 \\
 & 720 & \textbf{0.081} & {\ul 0.219} & \textbf{0.081} & \textbf{0.218} & 0.083 & 0.221 & 0.085 & 0.226 & 0.102 & 0.250 & 0.087 & 0.226 & 0.110 & 0.267 & 0.438 & 0.586 & 0.782 & 0.730 \\
 & \textbf{avg} & \textbf{0.052} & \textbf{0.172} & {\ul 0.054} & {\ul 0.174} & 0.055 & 0.175 & 0.056 & 0.178 & 0.069 & 0.202 & 0.056 & 0.179 & 0.081 & 0.221 & 0.281 & 0.441 & 0.523 & 0.536 \\
 \midrule
\multirow{5}{*}{\rotatebox{90}{ETTm2}} & 96 & \textbf{0.064} & \textbf{0.182} & 0.067 & {\ul 0.186} & 0.073 & 0.200 & 0.075 & 0.202 & 0.072 & 0.206 & 0.080 & 0.212 & {\ul 0.065} & 0.189 & 0.088 & 0.225 & 0.076 & 0.214 \\
 & 192 & \textbf{0.099} & \textbf{0.233} & {\ul 0.101} & {\ul 0.238} & 0.105 & 0.243 & 0.109 & 0.250 & 0.102 & 0.245 & 0.150 & 0.302 & 0.118 & 0.256 & 0.132 & 0.283 & 0.132 & 0.290 \\
 & 336 & \textbf{0.129} & \textbf{0.273} & 0.134 & {\ul 0.279} & 0.136 & 0.281 & 0.142 & 0.290 & {\ul 0.130} & {\ul 0.279} & 0.175 & 0.334 & 0.154 & 0.305 & 0.180 & 0.336 & 0.160 & 0.312 \\
 & 720 & 0.180 & {\ul 0.328} & 0.182 & 0.332 & 0.185 & 0.334 & 0.190 & 0.341 & {\ul 0.178} & \textbf{0.325} & 0.224 & 0.379 & 0.182 & 0.335 & 0.300 & 0.435 & \textbf{0.168} & 0.335 \\
 & \textbf{avg} & \textbf{0.118} & \textbf{0.254} & 0.121 & {\ul 0.259} & 0.125 & 0.265 & 0.129 & 0.271 & {\ul 0.121} & 0.264 & 0.157 & 0.307 & 0.130 & 0.271 & 0.175 & 0.320 & 0.134 & 0.288 \\
 \midrule
\multirow{5}{*}{\rotatebox{90}{ILI}} & 24 & {\ul 0.809} & {\ul 0.657} & 0.918 & 0.667 & 0.810 & 0.674 & 0.828 & 0.662 & \textbf{0.708} & \textbf{0.627} & 1.161 & 0.748 & 0.948 & 0.732 & 5.282 & 2.050 & 3.838 & 1.720 \\
 & 36 & 0.786 & 0.684 & 0.938 & 0.713 & 0.705 & {\ul 0.645} & 0.820 & 0.698 & \textbf{0.584} & \textbf{0.617} & 0.759 & 0.688 & {\ul 0.634} & 0.650 & 4.554 & 1.916 & 2.934 & 1.520 \\
 & 48 & \textbf{0.715} & \textbf{0.679} & 0.786 & 0.705 & 0.764 & 0.726 & 0.730 & {\ul 0.686} & {\ul 0.717} & 0.697 & 1.017 & 0.839 & 0.791 & 0.752 & 4.273 & 1.846 & 3.755 & 1.749 \\
 & 60 & \textbf{0.714} & \textbf{0.691} & {\ul 0.735} & {\ul 0.703} & 0.786 & 0.744 & 0.759 & 0.713 & 0.855 & 0.774 & 1.022 & 0.823 & 0.874 & 0.797 & 5.214 & 2.057 & 4.162 & 1.847 \\
 & \textbf{avg} & {\ul 0.756} & \textbf{0.678} & 0.844 & 0.697 & 0.766 & 0.697 & 0.784 & 0.690 & \textbf{0.716} & {\ul 0.679} & 0.990 & 0.775 & 0.812 & 0.733 & 4.831 & 1.967 & 3.672 & 1.709\\
 \bottomrule
\end{tabular}%
\vspace{-0.05in}
}
\label{tab:main2}

\end{table*}

\begin{table*}[t]
    \caption{Ablation study of \method on the Traffic dataset. MS stands for multi-scale feature and DT for dynamic tokenizer.}
\resizebox{0.9\textwidth}{!}{%
    \begin{tabular}{c|cc|cc|cc|cc|cc|cc|cc|cc}
    \toprule
        Model & \multicolumn{4}{c|}{Transformer} & \multicolumn{4}{c|}{Transformer+RoPE} & \multicolumn{4}{c|}{Transformer+MS+RoPE} & \multicolumn{4}{c}{Transformer+MS+gRoPE} \\
    \midrule
        Dynamic & \multicolumn{2}{c|}{w/o DT}  & \multicolumn{2}{c|}{w/ DT} &  \multicolumn{2}{c|}{w/o DT}  & \multicolumn{2}{c|}{w/ DT} & \multicolumn{2}{c|}{w/o DT} & \multicolumn{2}{c|}{w/ DT} & \multicolumn{2}{c|}{w/o DT} & \multicolumn{2}{c}{w/ DT} \\ 
    \midrule
        Metric & MSE & MAE & MSE & MAE & MSE & MAE & MSE & MAE & MSE & MAE & MSE & MAE & MSE & MAE & MSE & MAE \\ 
    \midrule
        96 & 0.426 & 0.270 & 0.424 & 0.271 & 0.420 & 0.270 & 0.415 & 0.270 & 0.421 & 0.270 & 0.416 & \textbf{0.266} & 0.418 & 0.268 & \textbf{0.414} & 0.267 \\ 
        192 & 0.439 & 0.276 & 0.432 & 0.272 & 0.432 & 0.274 & 0.431 & 0.274 & 0.432 & 0.274 & 0.432 & 0.272 & 0.429 & 0.272 & \textbf{0.427} & \textbf{0.271} \\ 
        336 & 0.452 & 0.283 & 0.447 & 0.280 & 0.445 & 0.280 & 0.444 & 0.278 & 0.448 & 0.281 & 0.442 & \textbf{0.277} & 0.445 & 0.281 & \textbf{0.440} & 0.278 \\ 
        720 & 0.487 & 0.301 & 0.483 & 0.297 & 0.481 & 0.300 & 0.478 & 0.298 & 0.484 & 0.301 & 0.477 & 0.297 & 0.478 & 0.298 & \textbf{0.474} & \textbf{0.296} \\ 
        \textbf{avg} & 0.451 & 0.283 & 0.447 & 0.280 & 0.445 & 0.281 & 0.442 & 0.280 & 0.446 & 0.282 & 0.442 & \textbf{0.278} & 0.443 & 0.280 & \textbf{0.439} & \textbf{0.278} \\ 
    \bottomrule
    \end{tabular}}
\label{tab:ablation}

\end{table*}

\subsection{Experimental Setting} 

\subsubsection{Dataset Description}

We conducted extensive experiments on time-series benchmark datasets. These datasets cover a variety of applications, including ETT and Electricity for electricity prediction, Exchange for financial applications, ILI for disease prediction, and Traffic for traffic prediction. 


\subsubsection{Baselines}
We compare \method with several Transformer-based models, including 
\begin{itemize}[leftmargin=3mm]
    \item \textbf{Reformer}~\cite{kitaev2020reformer}, which proposes a locality-sensitive hashing mechanism to reduce the time cost of self-attention calculation.
    \item \textbf{Informer}~\cite{zhou2021informer}, which proposes a ProbSparse self-attention with distilling techniques to extract the most important keys.
    \item \textbf{Autoformer}~\cite{Wu2021AutoformerDT}, which proposes an auto-correlation attention mechanism and a novel decomposition architecture.
    \item \textbf{FEDformer}~\cite{zhou2022fedformer}, which proposes to combine Fourier analysis with the Transformer-based method.
    \item \textbf{ETSformer}~\cite{Woo2022ETSformerES}, which exploits the principle of exponential smoothing and performs a layer-wise level, growth, and seasonal decomposition.
    \item \textbf{PatchTST}~\cite{nie2022time}, which divides the time-series data into sub-level patches to generate meaningful input features inspired by a patch-based Transformer on images in ~\cite{Dosovitskiy2020AnII}. 
\end{itemize}

\noindent We also consider several non-Transformer models, including
\begin{itemize}[leftmargin=3mm]
    \item \textbf{DLinear}~\cite{zeng2023transformers}, a simple linear model that only adopts a one-layer MLP model on the temporal dimension.
    \item \textbf{TimesNet}~\cite{wu2022timesnet}, which transforms the 1D time series into 2D space and extract the complex temporal variations from transformed 2D tensors by a parameter-efficient inception block.
    \item \textbf{Koopa}~\cite{liu2023koopa}, which tackles non-stationary time series with modern Koopman theory that fundamentally considers the underlying time-variant dynamics.
\end{itemize}

\subsubsection{Implementation details}

All the experiments are implemented with PyTorch ~\cite{paszke2019pytorch} and conducted on
a single NVIDIA RTX 3090 GPU. The hidden dimension size $D$ is set to $128$ for ETT datasets while $512$ for other datasets. We set the patch length $P$ to $16$ and the stride $S$ to $4$ for the ECL, Traffic, and ETT datasets while the patch length to $24$ and the stride to $2$ for ILI dataset. For dynamic linear, we set the number of groups $G$ to $8$ and the sparse ratio $SR$ to $0.5$. Assuming we have $i$ iterations for each epoch, we set the update frequency $\Delta t$ to $\lfloor 30\% \times i \rfloor$.
For the multi-scale transformer, we set $k$ for the number of multi-view sequences to 3 and $S_K=\{1,2,4\}$.

\begin{figure}[t]
    \centering
    \begin{minipage}[b]{0.5\textwidth}
    \centering
        \includegraphics[width=0.6\textwidth]{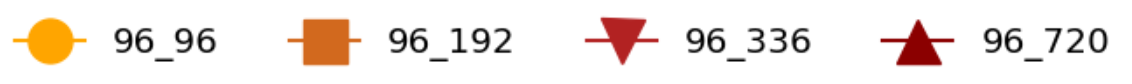}
    \end{minipage}
    \subfloat[ETTh1]{
        \centering
	\includegraphics[width=0.225\textwidth]{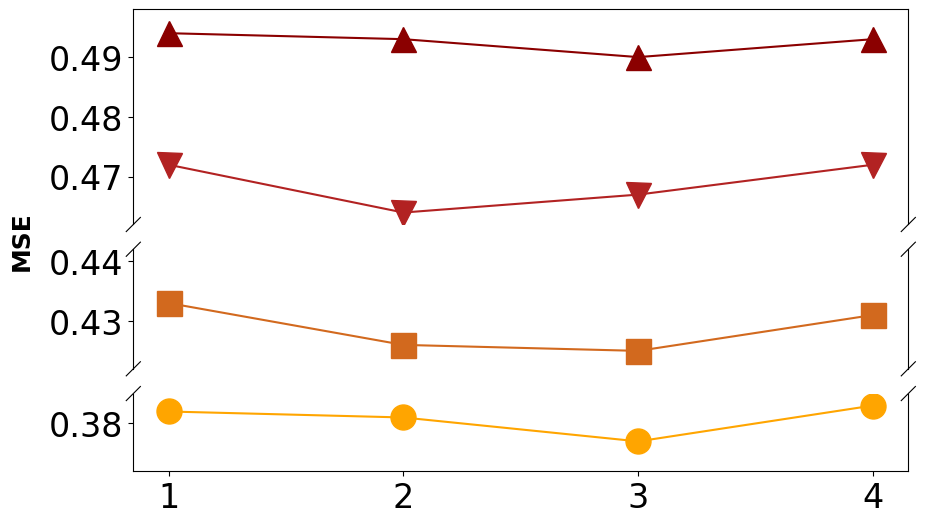}\label{fig:Traffic}
    }
    \subfloat[ETTm1]{
        \centering
	\includegraphics[width=0.225\textwidth]{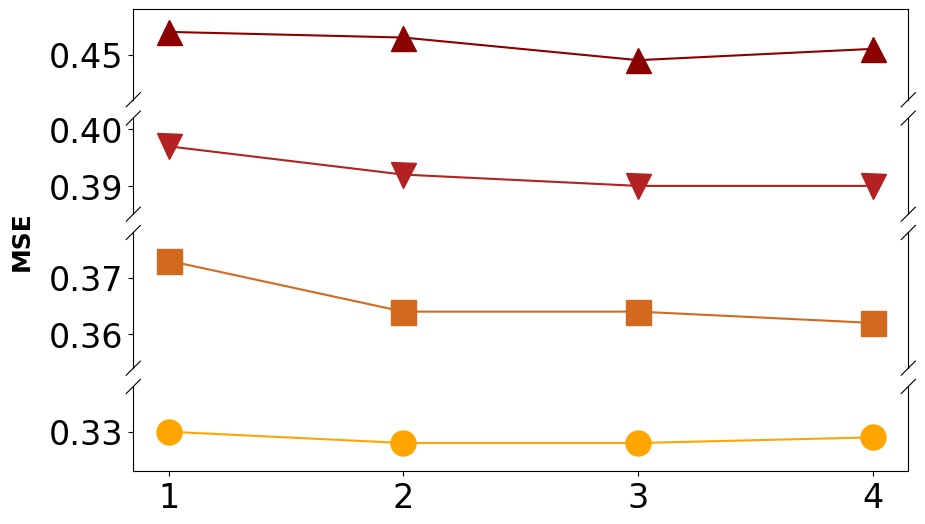}\label{fig:ETTh1}
    }
    \vspace{-0.1cm}
    \caption{The performance of DRFormer on ETTh1 and ETTm1 across varying numbers of multi-scale sequences.}
    \label{fig:parameter}

\end{figure}

\begin{figure}
    \centering
    \begin{minipage}[b]{0.5\textwidth}
    \centering
        \includegraphics[width=0.85\textwidth]{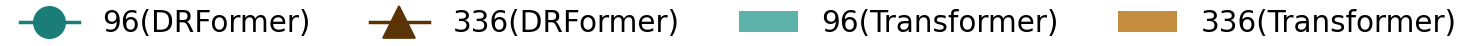}
    \end{minipage}
    \subfloat[ETTh2]{
        \centering
	\includegraphics[width=0.225\textwidth]{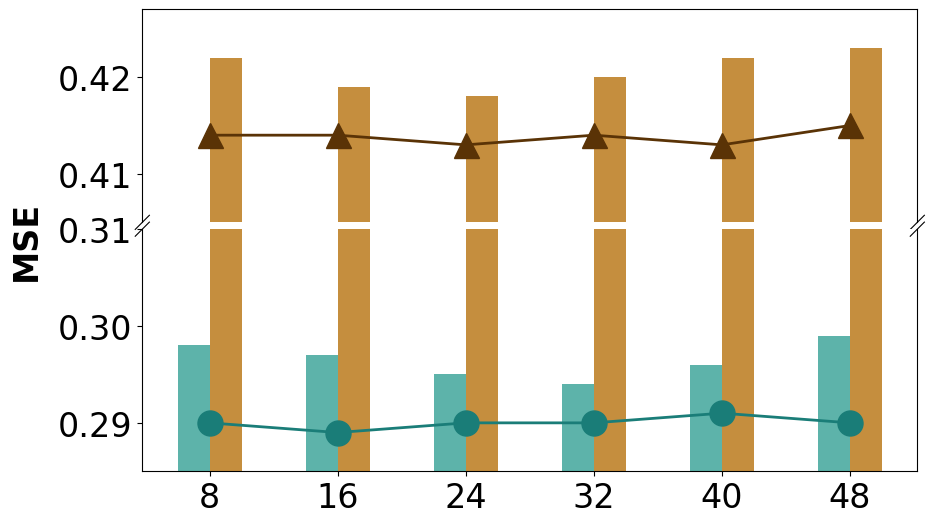}\label{fig:Traffic}
    }
    \subfloat[ETTm1]{
        \centering
	\includegraphics[width=0.225\textwidth]{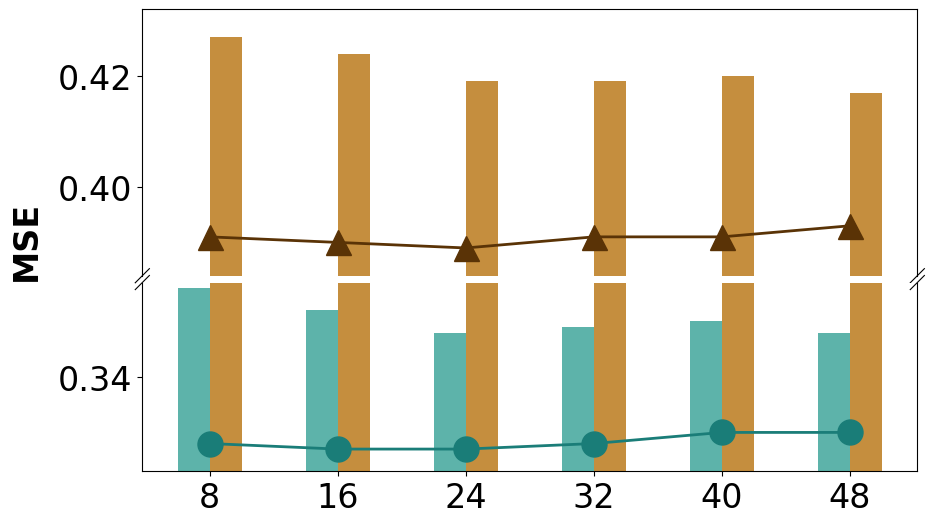}\label{fig:ETTh1}
    }
    \vspace{-0.1cm}
    \caption{The performance of Transformer(w/o DT) and DRFormer on ETTh2 and ETTm1 datasets under different predetermined patch lengths.}
    \label{fig:patch length}

\end{figure}

\begin{figure*}[htbp]
\centering
\begin{minipage}[b]{\textwidth}
\centering
    \includegraphics[width=0.45\textwidth]{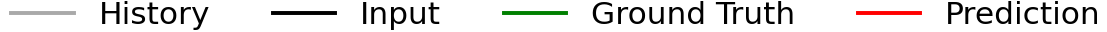}
\end{minipage}
\subfloat[DRFormer]{
        \centering
	\includegraphics[width=0.3\textwidth]{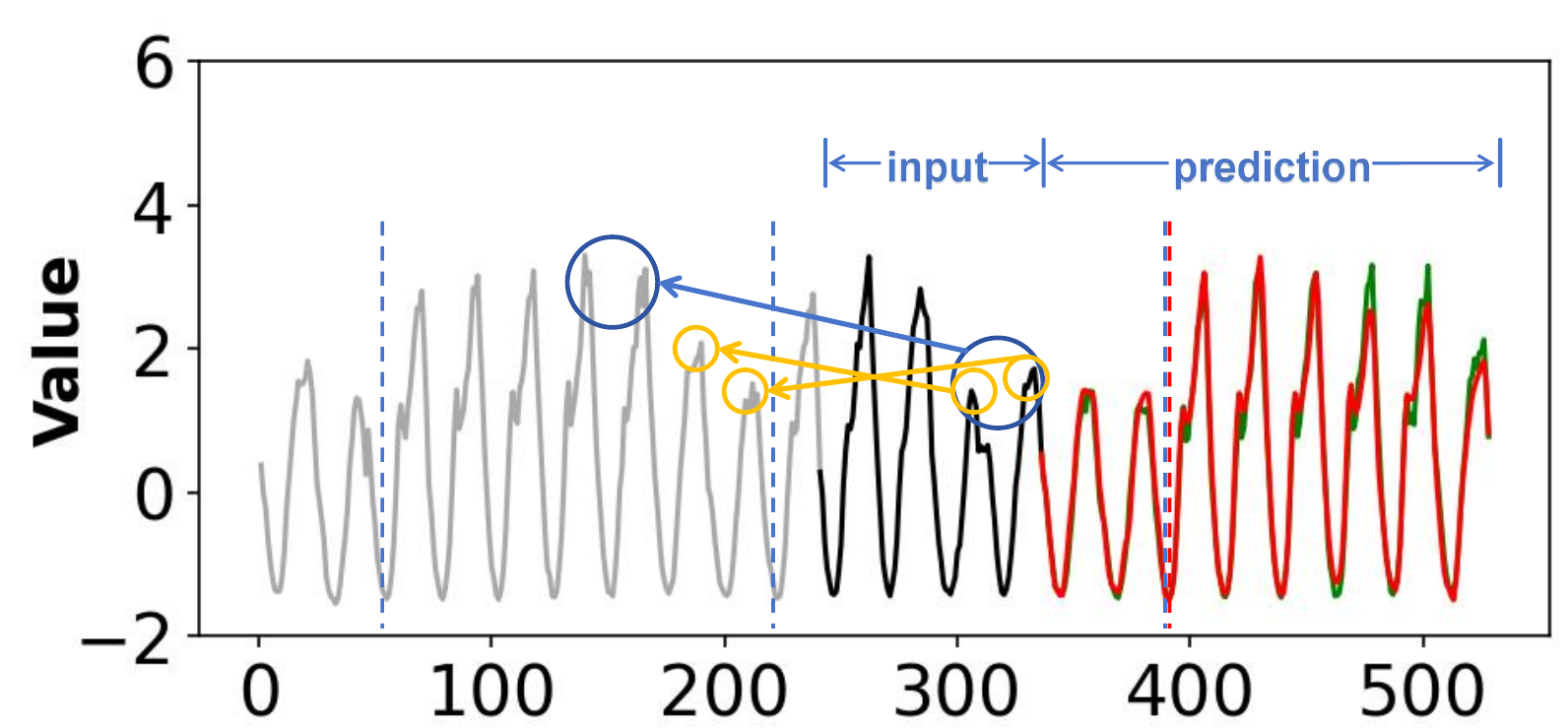}\label{fig:DRFormer}
}
\subfloat[PatchTST]{
        \centering
	\includegraphics[width=0.3\textwidth]{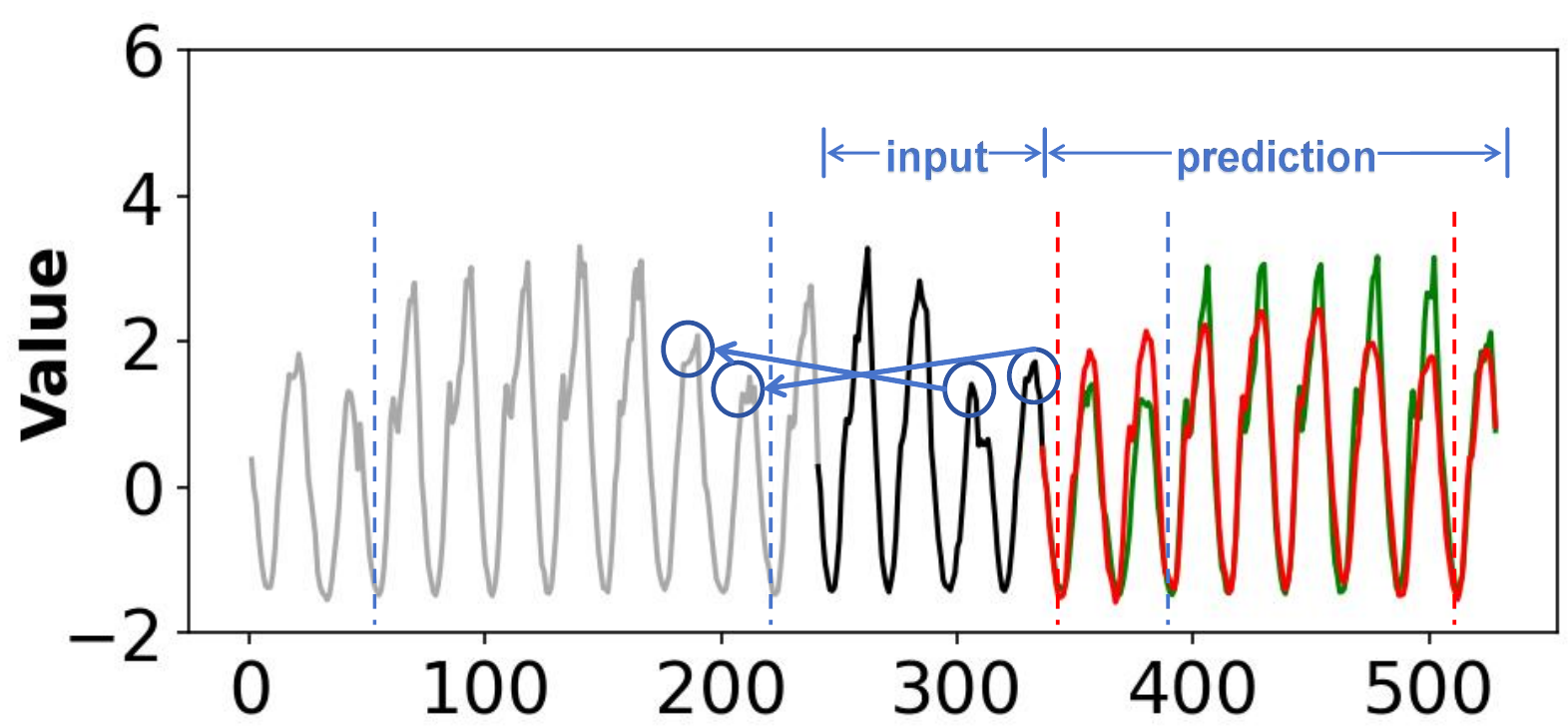}\label{fig:PatchTST}
}
\subfloat[Dlinear]{
        \centering
	\includegraphics[width=0.3\textwidth]{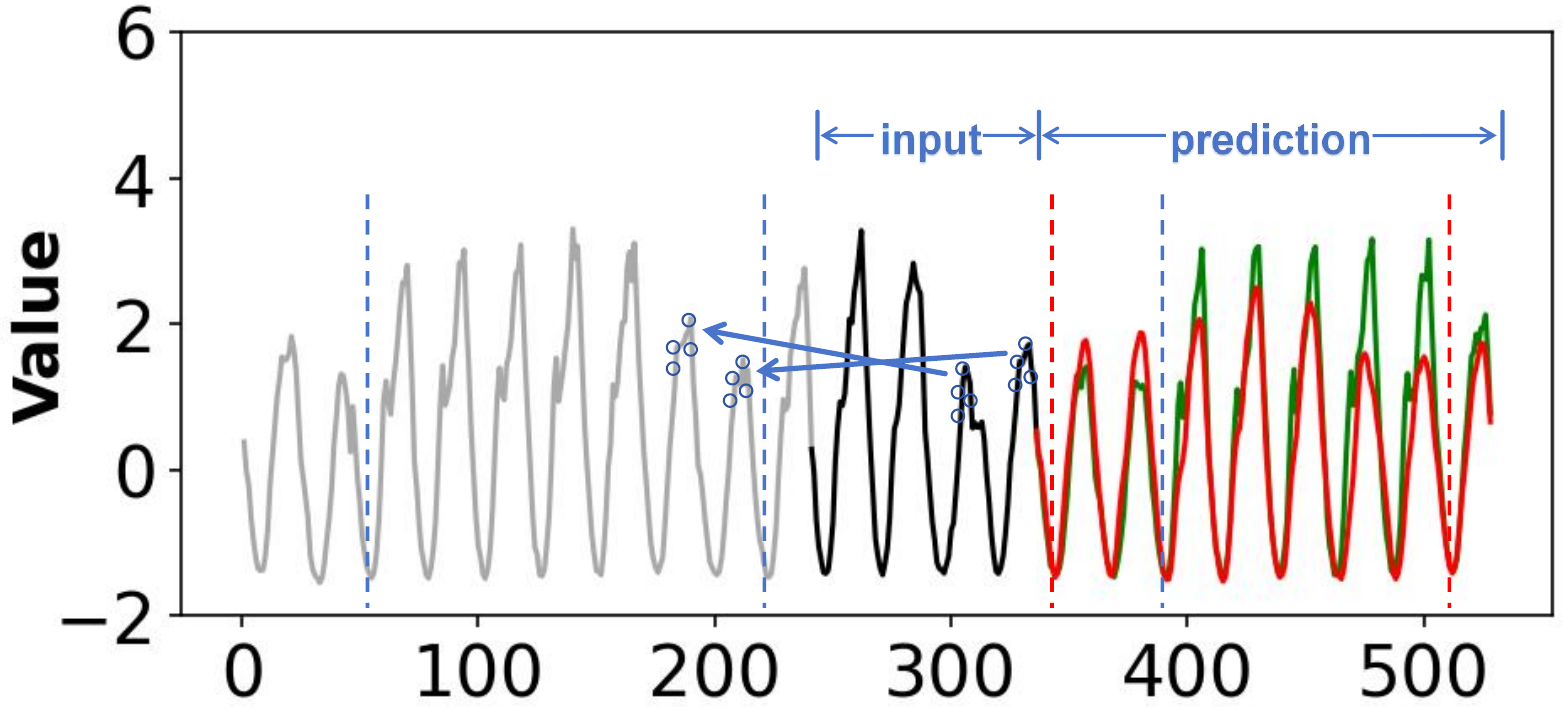}\label{fig:Dlinear}
}
\vspace{-0.9em}
\caption{Visualization of forecasting results on the Traffic dataset with I = 96 and O = 192. The black (grey) lines stand for input sequences (sequences before input). The green (red) lines stand for the ground truth (prediction). The blue (red) dashed lines represent the periodicity of the ground truth (prediction). Different diameters of circles represent different receptive fields.}
\label{fig:visualization}
\end{figure*}

\subsection{Model Comparisons}

\subsubsection{Multivariate Forecasting Results.}
The multivariate forecasting results are shown in Table~\ref{tab:main}, indicating \method achieves the state-of-the-art results on 6 datasets. Specifically, as compared to the best baselines, \method reduces the MSE by \textbf{8.4\%} (0.476 $\rightarrow$ \textbf{0.439}) on the Traffic dataset, and overall \textbf{2.7\%} (0.380 $\rightarrow$ \textbf{0.370}) reduction on four subsets of the ETT dataset, and \textbf{9.2\%} (2.139 $\rightarrow$ \textbf{1.959}) reduction on the ILI dataset. On average, \method achieves a \textbf{6.20\%} (0.617 $\rightarrow$ \textbf{0.581}) reduction on the MSE metric compared to the best baselines.
Additionally, compared with the best non-Transformer method, i.e., Koopa~\cite{liu2023koopa}, \method improves \textbf{18.22\%} on the Traffic dataset and an overall improvement of \textbf{8.26\%} under the MSE metric. Compared with the best Transformer-based method, i.e., PatchTST~\cite{nie2022time}, \method improves \textbf{16.54\%} on the ILI dataset and an overall improvement of \textbf{10.50\%}. 

\subsubsection{Univariate Forecasting Results.}
We show the univariate forecasting results ~\cite{Wu2021AutoformerDT} in Table ~\ref{tab:main2}. As shown in the table, \method achieves state-of-the-art results on four ETT datasets. Additionally, \method achieves the second-best results for the MAE metric on the ILI dataset. Specifically, \method achieves a \textbf{5.33\%} reduction on the ETTh1 dataset under the MSE metric, a \textbf{3.85\%} reduction on the ETTm1 dataset, a \textbf{2.63\%} reduction on the ETTh2 dataset and a \textbf{2.54\%} reduction on the ETTm2 dataset respectively.

\subsection{Ablation Study}
In this section, we delve into a comprehensive analysis of \method to showcase the effectiveness of each component of the model. 

\subsubsection{The effectiveness of dynamic modeling}
 We employ a dynamic tokenizer technique to capture fine-grained features within the patch size, which brings about diverse receptive fields. We demonstrate the effectiveness of the dynamic tokenizer using different Transformer architectures. As shown in Table \ref{tab:ablation}, the dynamic tokenizer can consistently decrease the prediction error, indicating the robustness of the dynamic tokenizer across various types of Transformer models.

\subsubsection{The effectiveness of multi-scale Transformer}
The consideration of multi-scale properties is a crucial aspect of time-series forecasting. To tackle this issue, we propose a hierarchical pooling strategy and a group-aware multi-scale Transformer model.
By comparing the results of Transformer+RoPE and Transformer+MS+gRoPE, as listed in Table ~\ref{tab:ablation}, we consider the design of multi-scale features effective when combined with group-aware RoPE. The comparison results between Transformer+RoPE and Transformer+MS+RoPE, with a few cases where performance decreases (MSE w/o DT), are reasonable. Transformer+MS+RoPE faces challenges in effectively aligning spatially close patches across varying scales and capturing intricate dependencies among different representation groups. In summary, the synergistic use of a multi-scale Transformer and gRoPE emerges as a requisite for optimal performance.

\subsubsection{The effectiveness of relative position embedding}

Incorporating relative position information in input sequences is crucial for Transformer-based models to overcome the weak sensitivity to the ordering of time series~\cite{zeng2023transformers}. 
To tackle the issue, we first apply RoPE on the Transformer model. 
By comparing the results of the Transformer and Transformer+RoPE, as listed in Table ~\ref{tab:ablation}, we can observe that RoPE improves the forecasting performance.

Furthermore, to overcome the limitations of position awareness of multi-scale representations for the transformer model, we propose a novel group-aware RoPE (gRoPE). By comparing the results of Transformer+MS+RoPE and Transformer+MS+gRoPE in Table ~\ref{tab:ablation}, we can observe that multi-scale Transformer models with gRoPE perform better than those with RoPE.

\subsection{Sensitivity Analysis}


In this section, we study the sensitivity of DRFormer to its hyperparameters and masking strategy. 



\subsubsection{The Influence of Multi-Scale Sequences}

Using multi-scale sequences allows us to extract features at multiple scales by transforming original resolution time series into multi-scale representations. To examine the impact of parameter $k$ on forecasting results, we varied $k$ within the range \{1, 2, 3, 4\} and evaluated the performance of \method in terms of mean squared error (MSE) on the ETTh1 and ETTm1 datasets. The results, depicted in Figure~\ref{fig:parameter}, demonstrate relatively stable and consistent trends across four different prediction horizons \{96, 192, 336, 720\}. Notably, increasing the value of $k$ leads to a significant reduction in MSE errors on both datasets, as long as $k$ remains below 3. This improvement can be attributed to the incorporation of features at more diverse scales through an increased number of multi-scale sequences. However, it is important to note that the length of the resized sequence decreases rapidly with larger kernel sizes, which ultimately limits the potential for further enhancement in forecasting performance.

    

\subsubsection{The Influence of Patch Length}

To analyze the impact of patch length on ETTh2 and ETTm1 datasets, we select patch length from $\{8,16,24,32,40,48\}$. Results from Figure ~\ref{fig:patch length} indicate that \method exhibits significant insensitivity to changes in patch length compared to patch-based Transformers without dynamic tokenizer and multi-scale sequences. The accuracy of \method on the test set remains consistently high across a wide range of patch-length configurations, highlighting the advantage of capturing a set of receptive fields with a predetermined patch length. 

\subsubsection{The Influence of Masking Strategy}

\begin{figure}[htbp]
    \centering
    \subfloat[Traffic]{
        \centering
	\includegraphics[width=0.225\textwidth]{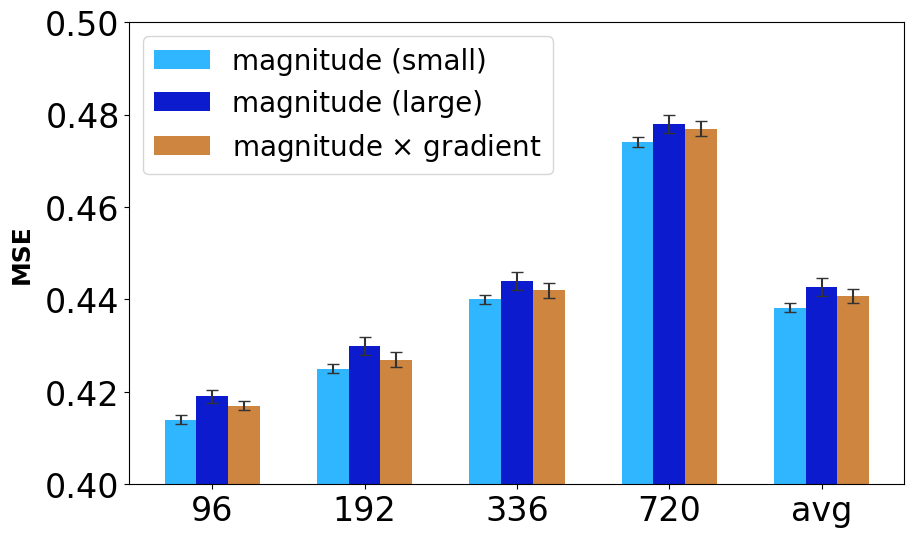}\label{fig:Traffic}
    }
    \subfloat[ETTh1]{
        \centering
	\includegraphics[width=0.225\textwidth]{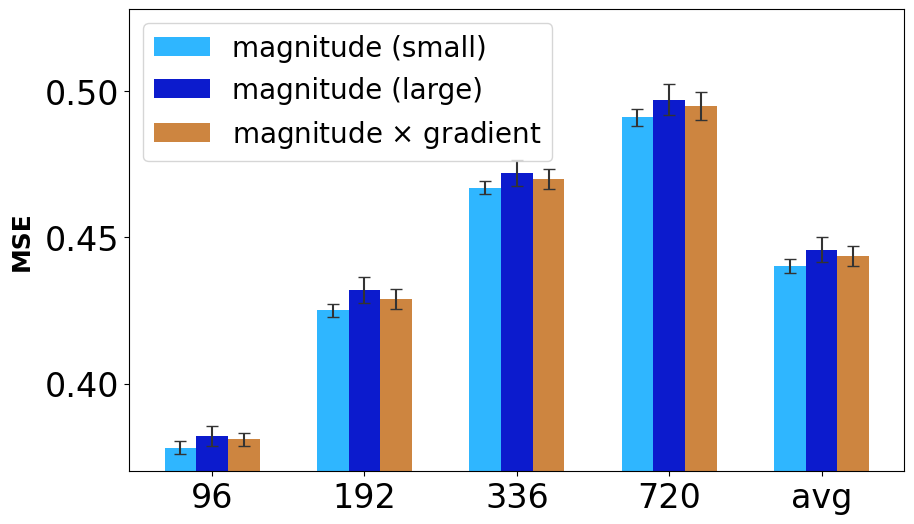}\label{fig:ETTh1}
    }
    \vspace{-0.2cm}
    \caption{The performance of DRFormer on Traffic and ETTh1 datasets across distinct masking strategies.}
    \label{fig:masking strategy}
\end{figure}

We explored various masking approaches, including masking out weights based on their magnitudes, both small and large, as well as masking weights according to the product of their magnitudes and gradients~\cite{Shrikumar2017LearningIF}.  We ultimately chose masking out weights with small magnitudes as it is intuitive and has been experimentally proven to be the most effective as shown in Figure ~\ref{fig:masking strategy}. It is widely recognized that the contribution of weights with smaller magnitudes is insignificant or even negligible.
\label{influence of strategy}

\begin{figure}[htbp]
    \centering
    \subfloat[Parameters]{
        \centering
	\includegraphics[width=0.225\textwidth]{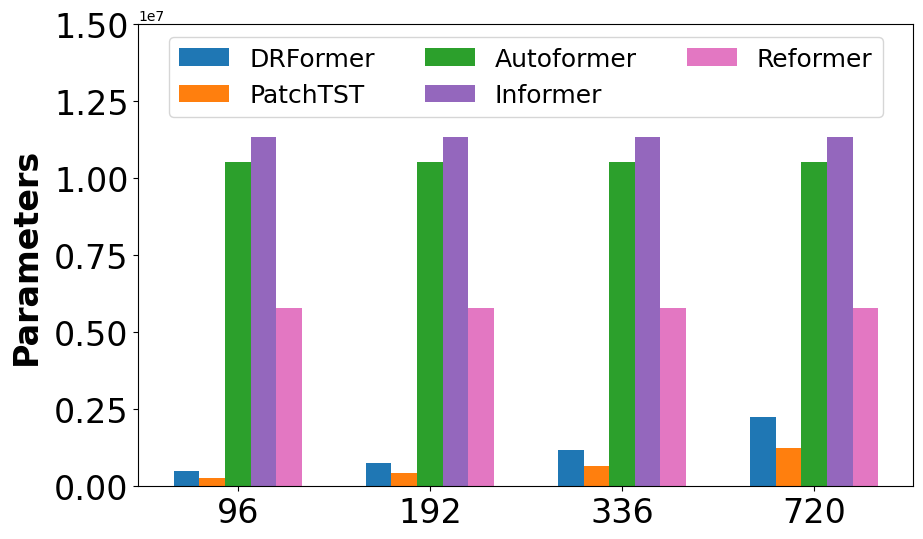}\label{fig:Parameters}
    }
    \subfloat[Training Time]{
        \centering
	\includegraphics[width=0.225\textwidth]{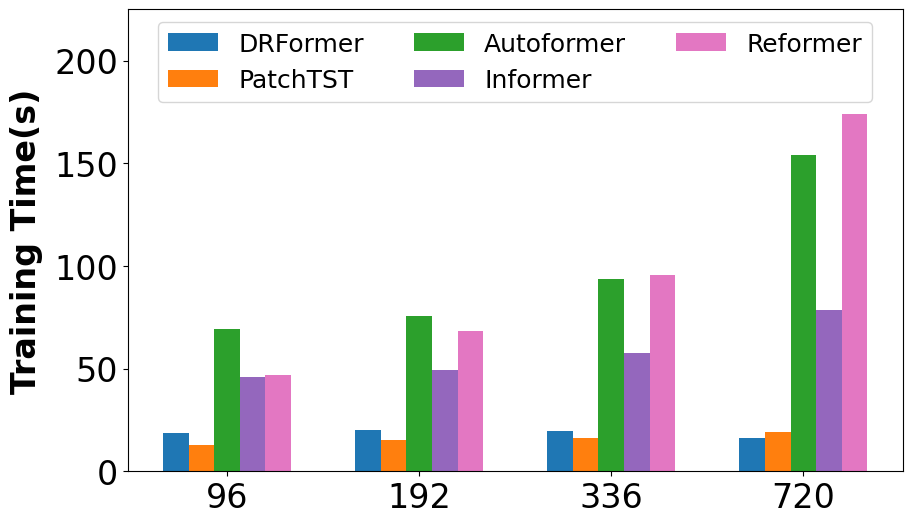}\label{fig:Training Time}
    }
    \vspace{-0.2cm}
    \caption{The comparison of parameters and training time between DRFormer and other transformer-based models.}
    \label{fig:complexity}

\end{figure}

\subsection{Model Complexity Analysis}
We conducted experiments to assess the complexity of DRFormer, focusing on two key metrics: parameters and training time. To ensure fairness, we maintained the same batch size for all models. As depicted in Figure ~\ref{fig:complexity}, DRFromer demonstrates significant advantages in both metrics, trailing only PatchTST. This can be attributed to the implementation of a multi-scale Transformer, which increases the total number of tokens by adding coarse-grained tokens via hierarchical max pooling. However, the additional resource requirements are deemed acceptable. In comparison to models Autoformer, Informer, and Reformer, DRFormer exhibits lower complexity.

\subsection{Visualization}
We select one test example from the Traffic dataset for case visualization. The ground truth and the predictions from \method and other baselines, i.e., PatchTST, and DLinear, are shown in Figure \ref{fig:visualization}, where \method provides the best forecasting. 
Specifically, we observe that \method, less affected by low amplitude at the end of the input sequence, relies on long-term trends to align accurately with corresponding segments in the historical sequence.
Compared with PatchTST and DLinear, the diverse receptive fields in Figure~\ref{fig:DRFormer} enable \method to learn multi-scale temporal patterns, improving its ability to predict periodicity and long-term variation without sacrificing compromising the accuracy of details.


\section{Conclusion}
In this paper, we propose a multi-scale Transformer model coupled with a dynamic tokenizer, named \method, for long-term time series forecasting. \method is a patch-based Transformer with a dynamic tokenizer and multi-resolution representations. Additionally, we present a novel group-aware RoPE method, named gRoPE to enhance intra- and inter-group position awareness among representations with different temporal scales. Extensive experimental results on both multivariate and univariate time series forecasting demonstrate that \method outperforms the previous state-of-the-art approaches.
Dynamic tokenizer and multi-scale Transformer can be transferred easily to other patch-based models.

\textbf{Limitations:} \method is designed under a channel-independent setting and it can be further explored to incorporate the correlation between different channels.



\section{Acknowledgments}
The authors would like to extend their heartfelt thanks to Zanwei Zhou and Jiajun Cui for their insightful feedback and valuable suggestions on paper writing. This work was supported in part by the National Key Research
and Development Program of China (2021ZD0111000, 2019YFB2102600) and Key Laboratory of Advanced Theory and Application in Statistics and Data Science, Ministry of Education.
\bibliographystyle{ACM-Reference-Format}
\bibliography{reference}

\appendix

\end{document}